\documentclass{article}

\usepackage{PRIMEarxiv}

\usepackage[utf8]{inputenc} 
\usepackage[T1]{fontenc}    
\usepackage{hyperref}       
\usepackage{url}            
\usepackage{booktabs}       
\usepackage{amsfonts}       
\usepackage{nicefrac}       
\usepackage{microtype}      
\usepackage{lipsum}
\usepackage{fancyhdr}       
\usepackage{graphicx}       
\usepackage{natbib}
\usepackage{tikz}
\usepackage{doi}
\usepackage{algorithm}
\usepackage{algpseudocode}
\usepackage{listings}
\usepackage{tabularx}
\title{Solving versus Verifying: Catching Contradictions in Tax Reasoning Systems}

\author{
  Albert Sadowski \\
  Faculty of Electronics and Information Technology \\
  Warsaw University of Technology \\
  Warsaw, Poland \\
  \texttt{albert.sadowski.stud@pw.edu.pl} \\
   \And
  Jarosław A. Chudziak \\
  Faculty of Electronics and Information Technology \\
  Warsaw University of Technology \\
  Warsaw, Poland \\
  \texttt{jaroslaw.chudziak@pw.edu.pl} \\
}

\begin{document}
\maketitle

\begin{abstract}
Large language models now compute correct tax liabilities on over 90\% of well-formed cases in statutory benchmarks, which makes them candidates for the tax-advisory and compliance systems that consume such an answer directly. Real legal inputs, however, are frequently defective: required facts are missing, or stated facts contradict one another. Accuracy on clean benchmarks says nothing about how a model behaves then, and a system that computes straight through a defective input returns a confident number with no sign that anything is wrong. This raises two questions: does a model asked to solve a case abstain when the input is defective, and when it does not, can the same model catch the defect when asked instead to verify the input? We study six recent models on SARA-derived tax cases under missing-fact and contradictory-fact perturbations. The strongest models abstain when a fact is missing but compute through injected contradictions, returning the clean-input answer 63-76\% of the time with no signal of the conflict; asked instead to verify the same input, they flag most of those contradictions. We wire that verification call into a simple contradiction gate: one extra call that abstains when the model reports a conflict. Across all six models it recovers most of the missed contradiction abstention at a clean-accuracy cost of at most about 5 percentage points, with no training and no external tooling. High accuracy on well-formed inputs is therefore an incomplete measure of reliability, and the detection the solver misses is cheaply recoverable with a single self-check.
\end{abstract}

\keywords{large language models \and legal reasoning \and contradiction detection \and self-verification \and tax calculation}

\section{Introduction}
\label{sec:intro}

Large language models (LLMs) have reached accuracy levels on statutory legal reasoning benchmarks that make deployment in legal advisory, tax preparation, and compliance workflows a realistic prospect. On the StAtutory Reasoning Assessment (SARA)~\cite{SARA} tax calculation task from LegalBench~\cite{legalBench}, recent models compute correct answers within $\pm$10\% on over 90\% of well-formed inputs. This level of performance has prompted growing interest in using LLMs as components of legal reasoning systems~\cite{matakChudziakGaius,sadowskiChudziakCIKM,lmsAndLogicProgramsForTaxReasoning}.

However, accuracy on well-formed inputs tells us little about how models behave when the input itself is defective. In real-world legal practice, inputs are routinely incomplete or inconsistent: clients omit required information, records conflict with each other, and documents contain contradictory statements~\cite{legalWiz}. A model that computes through such inputs poses a risk that is not captured by standard accuracy benchmarks.

Recent work has shown that LLMs struggle to abstain on unanswerable inputs in mathematics, general question answering, financial reasoning, and healthcare~\cite{abstentionBench,benchmarkingHallucinationInLLMs,realFin,whenSilenceIsSafer}, and that legal knowledge retrieval is prone to hallucination~\cite{largeLegalFictions}. Statutory tax calculation differs from these settings: it applies legal rules to facts in structured multi-step reasoning, where missing or contradictory information is routine~\cite{plawBench}. Existing abstention benchmarks primarily test underspecification, but real legal inputs also fail by contradiction, and whether models distinguish these failure modes remains unexplored.

\begin{figure}[t]
\centering
\includegraphics[width=0.6\columnwidth]{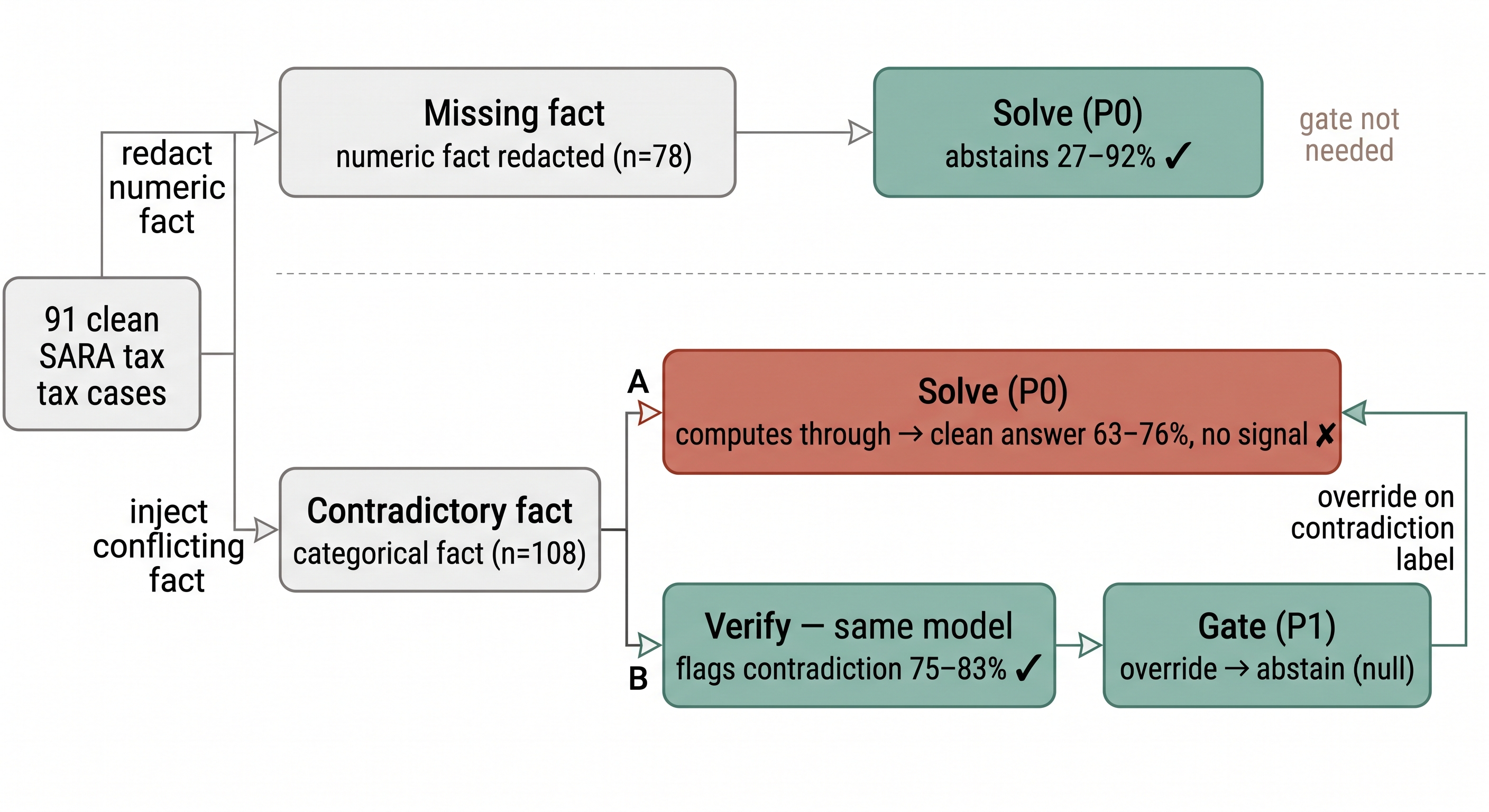}
\caption{Experimental design and the contradiction gate. The same model is asked to solve (return the tax liability or \texttt{null}) and to verify (flag missing or contradictory facts) on perturbed SARA cases; it abstains on missing facts but computes straight through injected contradictions (returning the clean answer 63-76\% of the time, top vs.\ bottom), while the same model asked to verify flags 75-83\% of them. The gate (P1) runs the verify call in parallel and overrides the solver to abstain on the contradiction label, recovering 70-83\% abstention at 0-5.1 percentage points of clean-accuracy cost (\S\ref{sec:gate}).}
\label{fig:overview}
\end{figure}

In this paper, we address two questions. First, can current LLMs recognise when a legal reasoning question cannot be reliably answered, either because required information is missing or because the input contains logical contradictions? Second, when a model fails to recognise a contradiction while solving, can the same model recognise it when asked instead to verify the input? We evaluate six recent LLMs on tax cases derived from the SARA benchmark~\cite{SARA} under three conditions (Fig.~\ref{fig:overview}): clean (unperturbed), redacted (a required numeric fact removed), and contradicted (a categorical fact supplemented with a conflicting value). We pair each defect type with the fact category where it appears clearly (\S\ref{sec:method}), and we evaluate the verification check from our second question entirely within the contradiction condition.

We make two contributions. The first is a diagnosis with a deployment consequence: high benchmark accuracy does not predict safe behaviour on inputs containing contradictory facts. On injected contradictions the four high-accuracy models abstain on at most 15\% while solving (three of the four under 5\%), and return the clean-input answer 63-76\% of the time instead, with no signal that the input was defective. Asked to verify the same inputs rather than solve them, the same models flag 75-83\% of those identical contradictions, so the detection is recoverable. As a calibration, on missing facts the same models almost never return the clean answer (preservation under 2\%), so the failure to abstain on contradictions is selective. The second is that this gap is cheaply actionable: a trivial one-call check, using the model itself as an input verifier and abstaining when it reports a contradiction (the \emph{contradiction gate}), recovers 70-83\% contradiction abstention at 0-5.1 percentage points of clean-accuracy cost, across all six models with no fine-tuning and no external tooling.
\section{Related Work}
\label{sec:related}

Our methodology draws on three established areas of research: LLM applications in legal reasoning, selective prediction (abstention), and how language models process contradictory information.

\subsection{LLMs for Legal Reasoning}

AI for legal reasoning dates back to expert systems for tax law~\cite{mccarty1977reflections}. The SARA dataset~\cite{SARA}, later included in LegalBench~\cite{legalBench}, benchmarks statutory reasoning over the US Internal Revenue Code; Blair-Stanek et al.~\cite{canGPT3PerformStatutoryReasoning} showed GPT-3 still made clear reasoning errors on it, especially on synthetic statutes.

Other benchmarks evaluate realistic practice scenarios~\cite{plawBench} and safety across professional domains~\cite{trident}, and Jurayj et al.~\cite{lmsAndLogicProgramsForTaxReasoning} combine LLMs with Prolog for tax reasoning. Prior work also explores multi-agent frameworks for verifiable legal reasoning~\cite{sadowskiChudziakCIKM} and structured prompting for rule application~\cite{sadowskiChudziakKES}. These show LLMs are accurate enough to warrant deployment discussions while highlighting persistent reasoning gaps.

\subsection{Abstention and Selective Prediction}

The ability of a model to recognise when it should not answer is related to the broader literature on selective prediction and uncertainty estimation. Kirichenko et al.~\cite{abstentionBench} introduce AbstentionBench, evaluating 20 frontier LLMs on unanswerable questions across 20 datasets. A key finding is that reasoning fine-tuning degrades abstention performance by 24\% on average, even in domains where reasoning models excel. Sun et al.~\cite{benchmarkingHallucinationInLLMs} confirm this pattern using unanswerable math word problems (UMWP), showing that LLMs tend to hallucinate answers rather than recognise unsolvable questions. Zhou et al.~\cite{whenSilenceIsGolden} show reinforcement learning can teach abstention, with limited out-of-distribution generalisation.

In specialised domains, Dai et al.~\cite{realFin} introduce RealFin, showing that general-purpose models tend to over-commit when financial questions have missing premises. Ren et al.~\cite{latentRefusal} detect unanswerable text-to-SQL queries from hidden-layer signals. Presacan et al.~\cite{whenSilenceIsSafer} review abstention in healthcare, distinguishing uncertainty-driven from safety-driven abstention and proposing a conceptual evaluation framework.

Other frameworks give formal guarantees through conformal, entropy-based, and ensemble methods~\cite{selectiveGenerationForControllableLanguageModels,multiLLMAdaptiveConformalInterface,tecp,conu,lmPolygraph,scope}. A simpler mechanism is to ask the model itself whether an input is answerable, which can expose a reliability signal the answering pass does not; we apply this to input validation and measure its effect directly.

These works focus on mathematics, general question answering, and knowledge retrieval; abstention in legal reasoning under information deficit has not been their focus.

\subsection{Contradiction Detection and Reasoning Under Conflict}

The question of how models handle contradictory information has been studied in several contexts. Hou et al.~\cite{wikiContradict} introduce WikiContradict, showing that models typically rely on one passage and disregard conflicting information when presented with contradictory evidence from Wikipedia. Mantravadi et al.~\cite{legalWiz} propose LegalWiz, a multi-agent framework for contradiction detection in legal documents, finding that cross-document contradictions remain a major challenge. These works study contradiction at the retrieval or document level, not at the level of input facts provided directly to a reasoning model.

Our work differs from these lines in that the problem lies in the input itself (missing or contradictory facts), not in the model's knowledge or retrieval pipeline, and we compare model behaviour across both defect types.

\section{Methodology}
\label{sec:method}

We construct a controlled evaluation by taking clean tax cases and applying two types of perturbation (redaction and contradiction), then measuring whether models detect the resulting defects. The dataset, perturbation pipeline, evaluation scripts, and all results are publicly available.\footnote{\url{https://doi.org/10.5281/zenodo.22329077}}

\subsection{Base Dataset}

We use 91 tax cases taken from the SARA (StAtutory Reasoning Assessment) subset of the LegalBench benchmark~\cite{legalBench,SARA}. SARA tests statutory reasoning over the US Internal Revenue Code; the subset included in LegalBench contains cases that have been standardised for benchmark use. Each case describes a taxpayer's situation (income, filing status, dependents, deductions) in natural language and asks for the tax liability, with ground-truth answers computed from the applicable statute.

\subsection{Perturbation Design}

We apply two types of perturbation, each corresponding to a way legal inputs fail in practice: required facts are missing, and stated facts contradict one another. We pair each perturbation type with the fact category for which it produces a clear, unambiguous defect: redaction targets numeric facts, contradiction targets categorical facts. The reverse pairings are less clean: a redacted categorical fact is often inferable from context, and two conflicting numeric values read as an update or an approximation rather than a clear contradiction.

Redaction removes a required numeric input, such as the taxpayer's income, making the question unanswerable: any dollar answer is unsupported. We generate 78 redacted variants across the 91 source cases; the drop from 91 reflects the manual-review filter below, which excludes cases where redaction still left the answer determined (e.g., only zero-valued fields were available). Each redaction targets a non-zero numeric fact the computation requires, and an LLM rewrites the text to omit the value without leaving placeholders. A model that recognises the gap should abstain.

Contradiction injects a contradictory statement about a categorical fact, creating a case where two logically incompatible claims coexist. After the validation described below, we retain 108 contradiction variants across three categories: marital status (49 cases), filing status (38 cases), and dependent count (21 cases). The variants come from 59 of the 91 source cases: a case yields one variant per eligible categorical fact, so a single case contributes between one and five variants. The contradictory sentence is inserted verbatim into the description at least one sentence away from the original fact, without hedging language. A model that recognises the conflict should abstain.

We generated the perturbations in three LLM-assisted phases: extracting and classifying facts (numeric vs.\ categorical), deterministically selecting which to perturb (non-zero numeric facts for redaction, categorical facts with a supported contrary mapping for contradiction), and rewriting the case text to apply the perturbation. All phases used GPT-5-nano\footnote{\url{https://developers.openai.com/api/docs/models/gpt-5-nano}} (OpenAI). The contradictory sentence itself comes from a fixed rule table (for example, \emph{files jointly} maps to \emph{files separately}); the LLM only rewrites the surrounding text. GPT-5-nano shares a provider with two evaluated models, GPT-5 mini and GPT-5.2. Because the injected sentence is rule-generated, inserted verbatim, and manually reviewed, we do not expect the generator to favour these two models.

Automated checks verified textual integrity for each perturbation (value absent, no placeholders, no length inflation, no hedging language, no prompt leakage, verbatim contradictory sentence).\footnote{Full criteria and validation code are available in the project repository.} Every case was then manually reviewed against fixed criteria, and cases that failed were removed; for contradictions, the criterion is that the injected sentence directly negates a stated fact, with both present and the conflict computationally relevant. A subsequent review removed 8 contradiction cases where the injected sentence did not produce a genuine conflict: consistent restatements (e.g., \emph{does not file jointly} alongside \emph{files separately}), statuses already entailed by the narrative (e.g., a widow or a completed divorce paired with \emph{not married}), and \emph{no dependents} injected where the only child belongs to a different taxpayer.

Table~\ref{tab:example} illustrates the three conditions for one case with GPT-5 mini's responses: it computes the correct \$6{,}621 on the clean input, abstains when the income is redacted, but ignores the injected ``file jointly''/``file separately'' contradiction and again returns \$6{,}621.

\begin{table}[t]
\caption{Example case under three conditions (question:
``How much tax does Alice have to pay in 2017?''; ground truth
\$6{,}621). Bold marks the perturbation.}
\label{tab:example}
\centering
\footnotesize
\setlength{\tabcolsep}{4pt}

\begin{tabularx}{\columnwidth}{@{}lXr@{}}
\toprule
Condition & Case description (abridged) & Output \\
\midrule
Clean & In 2017, Alice was paid \$39{,}212, and Bob had no income. Alice and Bob have been married since Feb 3rd, 2017. Alice and Bob file separately in 2017. & \$6{,}621 \\
\addlinespace
Redacted & In 2017, \textbf{Alice had income}, and Bob had no income. Alice and Bob have been married \ldots{} file separately in 2017. & \texttt{null} \\
\addlinespace
Contradicted & In 2017, Alice was paid \$39{,}212, and Bob had no income. \textbf{Alice and Bob file jointly in 2017.} Alice and Bob \ldots{} file separately in 2017. & \$6{,}621 \\
\bottomrule
\end{tabularx}
\end{table}

\subsection{Models Evaluated}

We evaluate six recent LLMs spanning four providers: GPT-5 mini and GPT-5.2 (OpenAI), Claude Sonnet 4.6 (Anthropic), Qwen3.7-Plus (Alibaba), Kimi K2.5 (Moonshot), and Gemini 2.5 Flash (Google).\footnote{Qwen3.7-Plus and Kimi K2.5 were accessed via Fireworks AI; the others via the providers' own APIs. Exact model identifiers are listed in the project repository.} Four of these are chain-of-thought (``thinking'') reasoners: GPT-5 mini, Kimi K2.5, Gemini 2.5 Flash, and Qwen3.7-Plus; GPT-5.2 and Claude Sonnet answer directly. This distinction matters because one might expect step-by-step reasoning to help on both clean-input accuracy and contradiction detection.

Models were queried through their providers' APIs with structured output (a JSON schema constraining the response to a single nullable numeric field) at temperature 0 where the API supported it. We query each model in its default configuration with no model-specific thinking parameters; for several models thinking cannot be disabled in any case, so the thinking-versus-direct split reflects defaults under a uniform setup. Each of the six models was evaluated three times on the full dataset to assess result stability.

\subsection{Evaluation Prompt and Response Classification}

Each model receives the applicable statute, the case description (clean, redacted, or contradicted), and the question, with instructions to act as an expert tax attorney and return the tax liability in USD, or \texttt{null} if the answer cannot be determined from the provided facts (full prompt in the project repository). We deliberately constrain the output to this binary choice rather than free text: the model has an explicit way to abstain, and hedging language cannot obscure whether it detected a problem. The trade-off is that we do not observe the model's reasoning trace, only its final decision. This simulates a deployment setting in which a downstream pipeline consumes the model's numeric output to compute the tax due.

We classify each response into three categories:
\begin{itemize}
    \item \textbf{Numeric (non-zero):} the model returned a specific dollar amount greater than zero.
    \item \textbf{Zero:} the model returned exactly \$0, which is a valid computed answer (some cases have zero tax liability).
    \item \textbf{Absent:} the model returned \texttt{null} or produced no parseable number. Only this category counts as abstention, since the prompt explicitly instructs the model to return \texttt{null} when it chooses to abstain.
\end{itemize}

\subsection{Metrics}

We measure model behaviour along three dimensions:
\begin{itemize}
    \item \textbf{Baseline accuracy:} the proportion of clean (unperturbed) inputs where the model's answer is within $\pm$10\% of the ground truth.
    \item \textbf{Abstention rate:} the proportion of perturbed inputs where the model returns an absent response.
    \item \textbf{Answer preservation:} the proportion of perturbed inputs where the model's answer is within $\pm$10\% of the clean-input ground truth. Cases where the model abstains count as non-preserved.
\end{itemize}

Abstention is the primary metric: it is the behaviour a defective input should trigger. Answer preservation is a diagnostic measure: high preservation means the model reproduced the clean answer despite the defect, the failure mode we measure. A low value is ambiguous on its own, so we read it alongside abstention; the remainder is \emph{wrong-different} (a number outside the $\pm$10\% band). For the high-accuracy models preservation reliably captures the model reproducing its clean answer; for GPT-5.2 and Claude Sonnet, whose clean answers are often wrong, it is less meaningful, so we instead compare error sizes on perturbed versus clean inputs.

\subsection{Self-Verification and the Contradiction Gate}
\label{sec:gate}

So far each model has been asked to \emph{solve}: to return the tax liability or abstain. We now add a second task and ask the model instead to \emph{verify} the input. In a separate call, the same model receives the same statute, case description, and question, but is asked only whether the facts are sufficient and internally consistent to answer the question, not to compute anything. The response is constrained to a single categorical label: \texttt{none} (the input is well-formed), \texttt{missing} (a required fact is absent), or \texttt{contradiction} (two facts conflict). As with the solver, we use structured output, so the verifier emits a decision rather than hedging prose.

We define two pipelines. \textbf{P0} is the solver alone: its output is whatever number or \texttt{null} the model returns. \textbf{P1} is the \emph{contradiction gate}: take the P0 output, but override it to abstention (\texttt{null}) whenever the verifier returns \texttt{contradiction}. The gate is deliberately contradiction-specific: it does not act on \texttt{missing}, because the solver already abstains on missing facts (\S\ref{sec:results}) and gating there would conflate the two effects. P1 adds exactly one model call to P0 and changes nothing on inputs the verifier passes. It adds no logic beyond overriding the answer on the \texttt{contradiction} label; we include it to test whether the detected conflict converts into abstention, not as a system we propose for deployment.

We report three quantities for the gate: the \textbf{detection rate} (the share of contradicted inputs the verifier labels \texttt{contradiction}), the \textbf{false-flag rate} (the share of clean inputs it labels \texttt{contradiction}, which drives the gate's cost), and the \textbf{clean-accuracy cost} (the difference in baseline $\pm$10\% accuracy on clean inputs with versus without the gate). The verifier is the same model as the solver: this holds the model fixed (same weights, solve versus verify), at the cost of correlated errors (\S\ref{sec:limitations}). Each of the six models is run through the verifier three times.

\section{Results}
\label{sec:results}

We first establish baseline accuracy on clean inputs, then examine model behaviour under redaction and contradiction. All results are means across three independent runs per model, with standard deviations reported where relevant. Three runs is a small sample for variance, so we read the standard deviations as rough indications of stability, not precise estimates.

\subsection{Baseline Accuracy on Clean Inputs}
\label{sec:baseline}

Table~\ref{tab:baseline} shows each model's accuracy on the 91 unperturbed cases. Four models cluster at the top at the $\pm$10\% tolerance: GPT-5 mini (96.7\%), Qwen3.7-Plus (93.4\%, $\sigma$=2.7\%), Gemini 2.5 Flash (91.6\%, $\sigma$=0.5\%), and Kimi K2.5 (89.4\%, $\sigma$=2.3\%). The other two score far lower: GPT-5.2 (29.7\%, $\sigma$=6.8\%) and Claude Sonnet (17.9\%, $\sigma$=1.9\%). We refer to the first four as ``high-accuracy models'' throughout, as they are well separated from the rest, and to GPT-5.2 and Claude Sonnet, which answer directly (\S\ref{sec:method}), as the ``direct-answer models''.

All four high-accuracy models reason step by step and both direct-answer models score far lower, so reasoning mode and clean-input accuracy coincide in our set. GPT-5 mini answered all 91 clean cases, while Claude Sonnet abstained on 16.5\%.

We include all six models, but the two contributions rest on different subsets. The diagnosis (the solve-versus-verify gap) is established on the four high-accuracy models, where preservation reliably reflects answer inertia. The two direct-answer models test whether the gate transfers to models that behave differently while solving; what the gate needs is only the change from P0 to P1 on a fixed model. On these two, verification raises contradiction abstention from 7.4\% and 43.2\% to 70.7\% and 78.4\% (\S\ref{sec:gate_results}). Their low clean accuracy may partly reflect the single-number schema, which leaves a model that does not reason internally little room to work through the arithmetic; it is not a parsing artifact, since they return plausible but wrong dollar amounts rather than nulls. The schema cannot explain the gate result, because the verification call asks for one consistency label, not a computation. We therefore interpret these two models only through the P0-to-P1 change and do not prompt-engineer to raise their baseline.

\begin{table}[t]
\caption{Baseline accuracy (\%) on 91 clean cases at several
tolerances. The $\pm$10\% column, our primary tolerance throughout
(answer preservation and the gate's cost use it), is the mean over
three runs; the other columns are from a single run and shown only
to convey the accuracy profile.}
\label{tab:baseline}
\centering
\footnotesize
\setlength{\tabcolsep}{4pt}

\begin{tabular}{@{}lrrrr@{}}
\toprule
Model & $\pm$1\% & $\pm$5\% & $\pm$10\% & $\pm$20\% \\
\midrule
GPT-5 mini        & 87.9 & 93.4 & \textbf{96.7} & 100.0 \\
Qwen3.7-Plus      & 90.1 & 93.4 & \textbf{93.4} & 98.9 \\
Gemini 2.5 Flash  & 82.4 & 87.9 & \textbf{91.6} & 94.5 \\
Kimi K2.5         & 81.3 & 89.0 & \textbf{89.4} & 96.7 \\
\midrule
GPT-5.2           &  6.6 & 20.9 & 29.7          & 50.5 \\
Claude Sonnet 4.6 &  4.4 & 12.1 & 17.9          & 31.9 \\
\bottomrule
\end{tabular}
\end{table}

\subsection{Contradiction: Models Do Not Report Inconsistency While Solving}

Before turning to contradictions, we confirm that abstention works at all. On redacted inputs (Table~\ref{tab:main_results}) no model returns the clean-input answer (preservation at most 1.3\%): models abstain on 27-92\% of them, or less safely emit an unsupported number or \$0. The contrast is starkest for Qwen3.7-Plus, the model most accurate at the tightest tolerance on clean inputs ($\pm$1\%, Table~\ref{tab:baseline}): it abstains on only 27.4\% of redacted inputs, so even for it the missing-versus-contradiction asymmetry is in preservation (0.9\% versus 63.3\%), not abstention. The models do abstain when a required fact is missing, so the contradiction behaviour we turn to next is selective, not general caution.\footnote{We treat \$0 as a computed output, not an abstention; \$0 rates on redacted inputs reach 47\% (Qwen3.7-Plus). A confident \$0 on an unanswerable input is as unsafe as any other wrong number.}

The high-accuracy models produce numeric answers on most contradictory inputs. Three of them rarely abstain: Kimi K2.5 never (0.0\%), GPT-5 mini 1.2\% ($\sigma$=1.2\%), and Gemini 2.5 Flash 4.6\% ($\sigma$=0.8\%); Qwen3.7-Plus is somewhat more cautious at 14.8\% ($\sigma$=3.0\%). All four nonetheless return the clean-input answer on the majority of contradictions: 76.2\% (Gemini), 74.7\% (GPT-5 mini), 67.6\% (Kimi), and 63.3\% (Qwen3.7-Plus). They largely ignore the contradiction and compute as if the conflicting statement were not present.

\begin{table}[t]
\caption{Abstention and answer-preservation rates (\%) on perturbed
inputs. ``Abstain'' denotes the percentage of inputs for which the
model returned no answer. ``Preserve'' denotes the percentage of
outputs matching the ground truth within $\pm$10\%.
Values are means across three runs ($\sigma \leq 4.2\%$, largest for
the direct-answer models).}
\label{tab:main_results}
\centering
\footnotesize
\setlength{\tabcolsep}{4pt}

\begin{tabular}{@{}l rr rr@{}}
\toprule
& \multicolumn{2}{c}{Redacted ($n$=78)}
& \multicolumn{2}{c}{Contradicted ($n$=108)} \\
\cmidrule(lr){2-3}
\cmidrule(lr){4-5}
Model & Abstain & Preserve & Abstain & Preserve \\
\midrule
GPT-5 mini        & 64.5 &  1.3 & \textbf{1.2}  & \textbf{74.7} \\
Qwen3.7-Plus      & 27.4 &  0.9 & \textbf{14.8} & \textbf{63.3} \\
Gemini 2.5 Flash  & 92.3 &  0.0 & \textbf{4.6}  & \textbf{76.2} \\
Kimi K2.5         & 67.5 &  1.3 & \textbf{0.0}  & \textbf{67.6} \\
\midrule
GPT-5.2           & 59.8 &  1.3 &  7.4 & 22.5 \\
Claude Sonnet 4.6 & 85.0 &  0.0 & 43.2 &  7.4 \\
\bottomrule
\end{tabular}
\end{table}

\begin{figure}[t]
\centering
\includegraphics[width=0.6\columnwidth]{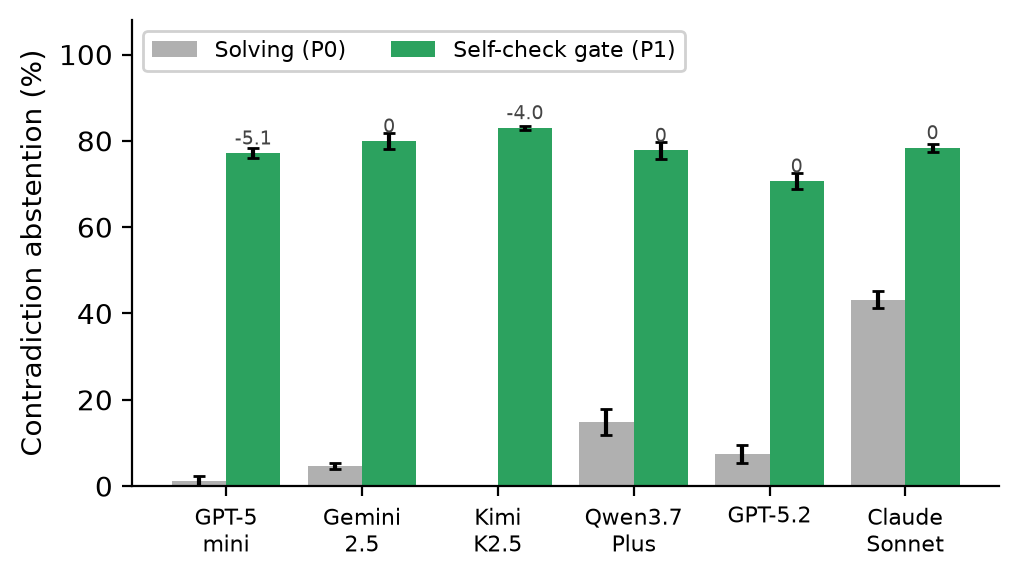}
\caption{Contradiction abstention while solving (P0) versus with the self-check gate (P1), per model. The P0-to-P1 jump is the abstention that solving leaves unused and the gate recovers; the label above each P1 bar is the clean-accuracy cost (percentage points). Error bars are the standard deviation across runs.}
\label{fig:abstention}
\end{figure}

The two direct-answer models are interpreted only for the gate, as noted above (\S\ref{sec:baseline}). Their error magnitudes on contradictions match their clean-input errors (Claude Sonnet: median 0.282 vs.\ 0.271; GPT-5.2: 0.218 vs.\ 0.189), so the abstention they show (Claude Sonnet 43.2\%, GPT-5.2 7.4\%) reflects general uncertainty rather than targeted detection.

Asked instead to verify the input (\S\ref{sec:gate}), the four high-accuracy models flag 75-83\% of these contradictions as inconsistent, well above the 0-15\% they abstain on while solving. The example from Table~\ref{tab:example} is typical: GPT-5 mini returns the clean-input \$6{,}621 on the contradicted case while solving, but the same model, asked to verify that same input, returns \texttt{contradiction} and names the conflict: ``the facts state both that Alice and Bob file jointly in 2017 and that they file separately in 2017.'' The conflict is legible to the model when the task is to check the input; the solving pipeline never asks. We apply this as a simple check next.

\subsection{The Contradiction Gate}
\label{sec:gate_results}

Table~\ref{tab:gate} reports the verifier and the gate. The detection column is the share of contradictions each model flags as \texttt{contradiction} when asked to verify. It is high and fairly uniform across capability tiers: 75-83\% for the four high-accuracy models and 68-77\% for the two direct-answer models. Set against solve-time abstention (Table~\ref{tab:main_results}: 0-15\% for the four high-accuracy models, 7.4\% and 43.2\% for the two direct-answer models), the verification step recovers a detection ability that solving leaves unused.

Applying P1 raises contradiction abstention to 70-83\% across all six models, from a P0 baseline as low as 0\% (Fig.~\ref{fig:abstention}). The answer-preservation rates that defined the diagnosis fall accordingly: for the high-accuracy models, the share of contradicted inputs answered with the clean-input value drops from 63-76\% to 11-15\%.

The false-flag rate (the share of clean inputs the verifier wrongly calls a contradiction) is 0.0-5.5\%. Among the capable models it is a real but modest clean-accuracy cost: 5.1 points for GPT-5 mini and 4.0 for Kimi K2.5, but essentially zero for Gemini 2.5 Flash and Qwen3.7-Plus (0.4). The two direct-answer models pay nothing measurable, partly because the clean cases they false-flag were already answered incorrectly.

The gate adds an inference call, so one might ask whether a second independent sample, rather than the verification step, is what helps. Treating a model's three solve runs as a self-consistency ensemble and abstaining when they disagree flags only 11-35\% of contradictions (11-21\% for the three strongest models, 35\% for Qwen3.7-Plus), well below the 75-83\% the verifier detects, and at a higher clean false-flag (3-14\% versus the gate's 0-5.5\%). This comparison has a limit: four of the six models run at temperature 0, so disagreement across their runs reflects nondeterminism rather than sampled diversity. Under our setup, resampling the solve task does not surface the contradiction, while asking the model to check the input does; a resampling baseline at nonzero temperature could behave differently, and we did not run one.

The verifier's labels are imperfect, but the gate tolerates this. Pooled across models, the verifier labels 77\% of contradicted inputs \texttt{contradiction}, 16\% \texttt{missing}, and 7\% \texttt{none}. On clean inputs it says \texttt{missing} 80\% of the time (60-98\% per model) and \texttt{none} only 17.5\%. This has a plain reading: SARA descriptions omit facts a strict reader can require, such as filing status, adjustments to income, itemised deductions, or age and blindness, and the verifier's justifications cite exactly these. The verifier is therefore not a general well-formedness judge. It works as a contradiction detector because the \texttt{contradiction} label is precise on clean inputs (2.5\% false flags) and the gate acts on that label alone. Its behaviour on the 78 redacted inputs is consistent with this: it says \texttt{missing} on 95-100\% of them per model and \texttt{contradiction} on only 0-4.7\%, so the \texttt{contradiction} label keeps its precision on defective inputs as well.
\begin{table}[t]
\caption{Self-verification and the contradiction gate on the 108
contradiction cases (means across three runs). Detect: contradictions
flagged \texttt{contradiction}. FalseFlag: clean inputs flagged
\texttt{contradiction}. Abstain (P1): contradiction abstention with the
gate, versus P0 in Table~\ref{tab:main_results}. Cost: drop in baseline
$\pm$10\% accuracy from the gate.}
\label{tab:gate}
\centering
\footnotesize
\setlength{\tabcolsep}{4pt}

\begin{tabular}{@{}lrrrr@{}}
\toprule
Model & Detect & FalseFlag & Abstain (P1) & Cost \\
\midrule
GPT-5 mini        & 77.2 & 5.5 & \textbf{77.2} & 5.1 \\
Qwen3.7-Plus      & 75.3 & 0.7 & \textbf{77.8} & 0.4 \\
Gemini 2.5 Flash  & 79.9 & 0.0 & \textbf{79.9} & 0.0 \\
Kimi K2.5         & 83.0 & 4.4 & \textbf{83.0} & 4.0 \\
\midrule
GPT-5.2           & 68.5 & 0.0 & 70.7 & 0.0 \\
Claude Sonnet 4.6 & 77.2 & 4.4 & 78.4 & 0.0 \\
\bottomrule
\end{tabular}
\end{table}

\subsection{Contradiction Type Analysis}

Detection is not uniform across contradiction types. Table~\ref{tab:contradiction_types} breaks the verifier's detection rate down by category for the high-accuracy models. Filing-status conflicts are caught almost always (95-97\%), marital-status conflicts moderately (78-83\%), and dependent-count conflicts much less often (30-59\%; only 6\% for the direct-answer GPT-5.2). The ordering matches logical hardness: \emph{file jointly} versus \emph{file separately} in the same year is a strict contradiction, whereas \emph{has a child} versus \emph{no dependents} is defeasible, since a child is a dependent only if it meets age and support tests. The lower dependent-count rate is therefore in part the verifier correctly declining to call a defeasible relation a strict contradiction, not only missed detection. The filing-versus-dependent gap is robust to sampling uncertainty (non-overlapping Wilson intervals), but the marital-versus-dependent ordering is not for Gemini and Kimi, where dependent-count detection is poorly constrained ($n$=21), so we draw no conclusion on that pair. Within each type the variants come from distinct source cases, except for one duplicated case in marital status and one in filing status, so the per-type intervals are not distorted by sibling variants of one case.

\begin{table}[t]
\caption{Verifier detection rate (\%) by contradiction type for
high-accuracy models (Gemini = Gemini 2.5 Flash, Kimi = Kimi K2.5,
Qwen = Qwen3.7-Plus), means across three runs. Wilson 95\% intervals
over the $n$ items are wide for dependent count (e.g., GPT-5 mini
32\% [16,53]). Detection is the percentage of contradicted inputs
flagged \texttt{contradiction}.}
\label{tab:contradiction_types}
\centering
\footnotesize
\setlength{\tabcolsep}{4pt}

\begin{tabular}{@{}l rrrr@{}}
\toprule
Contradiction type
& GPT-5 mini
& Gemini
& Kimi
& Qwen \\
\midrule
Marital status ($n$=49)  & 83.0 & 78.9 & 82.3 & 77.6 \\
Filing status ($n$=38)   & 94.7 & 94.7 & 97.4 & 97.4 \\
Dependent count ($n$=21) & 31.7 & 55.6 & 58.7 & 30.2 \\
\bottomrule
\end{tabular}
\end{table}

Dependent-count conflicts are the hardest to flag even when asked, because the injected claim is often not a strict contradiction. The residual misses concentrate here, and we do not read them all as detection failures; many are defeasible relations rather than strict conflicts, which is also why the clearest misfires were removed from the set (\S\ref{sec:method}). This reading does not conflict with the retention criterion in \S\ref{sec:method}. That criterion is textual: the injected sentence directly negates a stated fact, as \emph{no dependents} negates \emph{has a child}. Defeasibility concerns the legal relation between the facts: the child counts as a dependent only if age, residency, and support tests are met. A case can meet the textual criterion and still be legally defeasible. The verifier's fallback labels do not settle how to read the misses. On the undetected dependent-count cases it returns \texttt{missing} more often than \texttt{none}, but \texttt{missing} is also its dominant label on clean inputs (\S\ref{sec:gate_results}), so the label distribution carries little evidence either way.
\section{Discussion}
\label{sec:discussion}

For a deployment pipeline the practical finding is simple: asked to solve, the model computes through an injected contradiction and emits a confident answer; asked to check the same input, the model reports the conflict. We do not settle why solving misses it. One possibility: redaction creates a \emph{gap}, a step with no input, so the model cannot proceed, while contradiction provides \emph{two competing inputs}, so the model has something to work with at every step and resolves the conflict on its own, likely by defaulting to one value. Whatever the mechanism, the verification step recovers the missed abstention, which is what the gate exploits.
\subsection{Implications for Legal AI Deployment}

High benchmark accuracy alone is not sufficient evidence that a model will behave safely on real-world legal inputs, where contradictory client information is common. This adds to documented reliability concerns in legal contexts~\cite{largeLegalFictions,trident,plawBench}: the model may fail to recognise the task should not be solved at all, even when it could solve a clean version correctly.

The missed detection is easy to recover: the gate is one self-verification call, with no training and no external solver (Table~\ref{tab:gate}). It is the simplest of a range of options: upstream input checks~\cite{legalWiz} and output-side checks such as multi-agent disagreement, conformal ensembles~\cite{multiLLMAdaptiveConformalInterface,conu,tecp,scope}, and neuro-symbolic verification~\cite{sadowskiChudziakCIKM,sadowskiChudziakKES,Kant2025,Pan2024,Calanzone2024}. These are heavier but catch the defeasible conflicts the gate misses (\S\ref{sec:limitations}); we see it as a cheap first check, not a replacement.

\section{Evaluation and Future Work}
\label{sec:limitations}

Our findings connect to the three lines of work in \S\ref{sec:related}. The abstention literature reports that reasoning-tuned models often abstain less rather than more~\cite{abstentionBench}; in our set the step-by-step reasoners are the models that preserve the clean-input answer on contradicted inputs (the direct-answer models also return numbers, but do not reproduce the clean value), consistent with that pattern, though we cannot separate reasoning mode from capability, so we read it as agreement rather than confirmation. Work on unanswerable questions studies underspecified or missing-premise inputs~\cite{benchmarkingHallucinationInLLMs,realFin}; our models reproduce that behaviour on missing facts, but a contradiction is over-specified rather than under-specified, and the same models do not catch it, so contradiction is a distinct failure mode that underspecification benchmarks do not isolate. Finally, computing through a conflict by defaulting to one value is the input-level analogue of what WikiContradict~\cite{wikiContradict} found at the retrieval level, where models rely on one passage and disregard the conflicting one. Against this background our addition is the solve-versus-verify gap and the same-model gate that turns it into safe behaviour.

This result comes with several bounds. It is narrow in scope: the SARA-derived cases are stylised compared to real-world tax inputs, which involve natural-language narratives with more ambiguity and complexity, and they cover a single legal domain (US federal income tax) and a single benchmark. Whether the missing-vs-contradictory asymmetry generalises to other legal settings (contract analysis, regulatory compliance, criminal procedure) is an open question; the SARA-specific scope is a deliberate choice for this study rather than a claim about all legal AI. Since each defect type is paired with one fact category (\S\ref{sec:method}), the asymmetry is also measured across fact categories, and we do not disentangle the two.

The verifier, too, is a deliberate choice: the same model and weights as the solver, so shared blind spots can let a contradiction pass both stages. Yet it is the right design for the diagnostic claim: it shows that the same weights that compute through the conflict report it when the task changes, with no external knowledge added. A cross-model or external verifier answers a complementary question, whether the detection is recoverable from a different model. A specific external verifier can of course do worse than the same-model gate. But the same-model verifier remains one of the available choices, so the best gate over verifier choices is at least what we report; in that sense our numbers are a lower bound on what verification gating can recover.

Detection is weakest on dependent-count contradictions (6-59\%), many of which are defeasible rather than strict conflicts; where strict treatment is wanted, a stronger consistency check would be needed.

These limits leave several directions open. The methodology and gate should be tested in other legal domains, to see whether the effect and the check transfer beyond tax. The diagnosis can be sharpened: logging reasoning traces would show whether the solver detects the conflict internally. A cross-model or external verifier would test whether detection is recoverable from different weights, and stronger verifiers, neuro-symbolic~\cite{Calanzone2024,Kant2025,Pan2024} or conformal and ensemble methods~\cite{multiLLMAdaptiveConformalInterface,conu,tecp,lmPolygraph,scope,selectiveGenerationForControllableLanguageModels}, could catch the defeasible conflicts the same-model gate misses.

\section{Conclusion}
\label{sec:conclusion}

Large language models now solve well-formed statutory tax cases accurately enough to be considered for advisory and compliance systems, but real inputs are often defective: required facts are missing, or stated facts conflict. We evaluated six recent models on SARA-derived tax cases under both defects and found opposite behaviour. When a required fact is missing, the models abstain, or at least almost never reproduce the clean answer. When the input contradicts itself, the strongest models compute straight through the conflict and return the answer they would give to the clean case, with no signal that anything is wrong. The detection ability is there: asked to verify the same input rather than solve it, the same models report the conflict.

This solve-versus-verify gap gives a diagnosis and a simple check that acts on it. The diagnosis is that high accuracy on well-formed inputs does not predict safe behaviour on contradictory ones, so benchmark accuracy alone is an incomplete measure of reliability for legal reasoning. The check is the contradiction gate: a single self-verification call that reuses the same model, with no training and no external tooling, and it converts the recovered detection into abstention across all six models at little clean-accuracy cost. The gate is an easy first check, not a complete solution; extending it to other legal domains and to stronger verifiers that catch defeasible conflicts (\S\ref{sec:limitations}) is future work.

\bibliographystyle{unsrtnat}
\bibliography{references}

\end{document}